\def\year{2019}\relax
\documentclass[letterpaper]{article} %
\usepackage{aaai19}  %
\usepackage{times}  %
\usepackage{helvet} %
\usepackage{courier}  %
\usepackage[hyphens]{url}  %
\usepackage{graphicx} %
\usepackage{subfig}
\usepackage{graphicx}  %
\usepackage{times}

\usepackage{soul}
\usepackage{url}
\usepackage[utf8]{inputenc}
\usepackage{graphicx}
\usepackage{amsmath}
\usepackage{booktabs}
\usepackage{algorithm}
\usepackage{algorithmic}
\usepackage{xspace} 
\usepackage{amssymb}
\usepackage{bm}
\usepackage{color}
\usepackage{url}
\usepackage{booktabs}

\title{Action Guidance with MCTS for Deep Reinforcement Learning} %
\author{Bilal Kartal,\thanks{Equal contribution} Pablo Hernandez-Leal$^*$ and Matthew E. Taylor\\
Borealis AI\\
Edmonton, Canada\\
\{bilal.kartal, pablo.hernandez, matthew.taylor\}@borealisai.com
}
\begin{document}

\maketitle

\begin{abstract} 

Deep reinforcement learning has achieved great successes in recent years, however, one main challenge is the sample inefficiency. In this paper, we focus on how to use action guidance by means of a non-expert demonstrator to improve sample efficiency in a domain with sparse, delayed, and possibly deceptive rewards: the recently-proposed multi-agent benchmark of Pommerman. We propose a new framework where even a non-expert simulated demonstrator, e.g., planning algorithms such as Monte Carlo tree search with a small number rollouts, can be integrated within asynchronous distributed deep reinforcement learning methods. Compared to a vanilla deep RL algorithm, our proposed methods both learn faster and converge to better policies on a two-player mini version of the Pommerman game.

\end{abstract}

\maketitle

\section{Introduction}

\noindent Deep reinforcement learning (DRL) has enabled better scalability and generalization for challenging domains~\cite{arulkumaran2017deep,li2017deep,hernandez2018multiagent} such as Atari games~\cite{mnih2015human}, Go~\cite{silver2016mastering} and multiagent games (e.g., Starcraft II and DOTA 2)~\cite{openfive}. However, one of the current biggest challenges for DRL is sample efficiency~\cite{yu2018towards}.

On the one hand, once a DRL agent is trained, it can be deployed to act in real-time by only performing an inference through the trained model. On the other hand, planning methods such as Monte Carlo tree search (MCTS)~\cite{browne2012survey} do not have a training phase, but they perform computationally costly simulation based rollouts (assuming access to a simulator) to find the best action to take. %

There are several ways to get the best of both DRL and search methods. AlphaGo Zero~\cite{silver2017mastering} and Expert Iteration~\cite{anthony2017thinking} concurrently proposed the idea of combining DRL and MCTS in an imitation learning framework where both components improve each other. The idea is to combine search and neural networks \emph{sequentially} in a loop. First, search is used to generate an expert move dataset, which is used to train a policy network~\cite{guo2014deep}. Second, this network is used to improve expert search quality~\cite{anthony2017thinking}. These two steps are repeated. However, expert data collection by search algorithms can be slow in a sequential framework depending on the simulator~\cite{guo2014deep}, which can render the training too slow. 

In this paper, complementary to the aforementioned existing work, we show it is also possible to blend search with distributed model-free DRL methods such that search and neural network components can be executed \emph{simultaneously} in an \textit{on-policy} fashion. The main focus of this work is to show how to use relatively weak demonstrators (i.e., lightweight MCTS with a small number of rollouts) for model-free RL by coupling the demonstrator and model-free RL through an auxiliary task~\cite{jaderberg2016reinforcement}. The idea is that this auxiliary task would provide denser signals and more efficient learning with better exploration.

In this paper, we experiment on the recently proposed benchmark for (multi-agent) reinforcement learning: Pommerman~\cite{resnick2018pommerman}. This environment is based on the classic console game \emph{Bomberman}. Pommerman is  challenging game for many standard RL algorithms due its multiagent nature and its delayed, sparse, and deceptive rewards.

\begin{figure}
\centering
\includegraphics[scale=0.26]{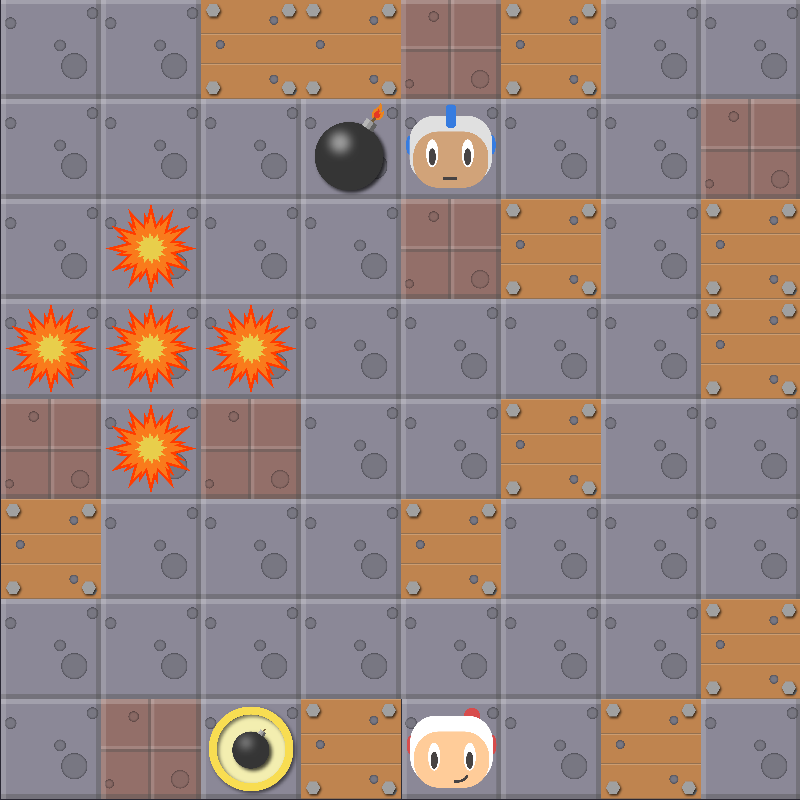}
\caption{An example of the $8 \times 8$ Mini-Pommerman board, randomly generated by the simulator. Agents' initial positions are randomized among four corners at each episode.}
\label{fig:pom8x8}
\end{figure}

We consider Asynchronous Advantage Actor-Critic (A3C)~\cite{mnih2016asynchronous} as a baseline algorithm and we propose a new framework based on diversifying some of the workers of A3C with MCTS-based planners (serving as non-expert demonstrators) by using the parallelized asynchronous training architecture. This has the effect of providing action guidance, which in turns improves the training efficiency measured by higher rewards.

\section{Related Work}  

Our work lies at the intersection of different areas, here we review the related existing work.

\textbf{Safe RL:} Safe Reinforcement Learning tries to ensure reasonable system performance and/or respect safety constraints during the learning and/or deployment processes~\cite{garcia2015comprehensive}. Roughly, there are two ways of doing safe RL: some methods adapt the optimality criterion, while others adapt the exploration mechanism. Our work uses continuous action guidance of lookahead search with MCTS for better exploration. 

\textbf{Imitation Learning:} Domains where rewards are delayed and sparse are difficult exploration problems when learning \textit{tabula rasa}. Imitation learning can be used to train agents much faster compared to learning from scratch. 

Approaches such as DAgger~\cite{ross2011reduction} formulate imitation learning as a supervised problem where the aim is to match the demonstrator performance. However, the performance of agents using these methods is upper-bounded by the demonstrator. Recent works such as Expert Iteration~\cite{anthony2017thinking} and AlphaGo Zero~\cite{silver2017mastering} extend imitation learning to the RL setting where the demonstrator is also continuously improved during training. There has been a growing body of work on imitation learning where demonstrators' data is used to speed up policy learning in RL~\cite{hester2017deep}. 

Hester et al.~(\citeyear{hester2017deep}) used demonstrator data by combining the supervised learning loss with the Q-learning loss within the DQN algorithm to pre-train and showed that their method achieves good results on Atari games by using a few minutes of game-play data. Kim et al.~(\citeyear{kim2013learning}) proposed a learning from demonstration approach where limited demonstrator data is used to impose constraints on the policy iteration phase. Another recent work~\cite{bejjani2018planning} used planner demonstrations to learn a value function, which was then further refined with RL and a short-horizon planner for robotic manipulation tasks.

\textbf{Combining Planning and RL:} Previous work~\cite{leonetti2016synthesis} combined planning and RL in such a way that RL can explore on the action space filtered down by the planner, outperforming using either solely a planner or RL. Other work~\cite{lee2019wisemove} employed MCTS as a high-level planner, which is fed a set of low-level offline learned DRL policies and refines them for safer execution within a simulated autonomous driving domain. A recent work by Vodopivec et al.~(\citeyear{vodopivec2017monte}) unified RL, planning, and search. 

\textbf{Combining Planning, RL, and Imitation Learning:} AlphaGo~\cite{silver2016mastering} defeated one of the strongest human Go players in the world. It uses imitation learning by pretraining RL's policy network from human expert games with supervised learning~\cite{lecun2015deep}. Then, its policy and value networks keep improving by self-play games via DRL. Finally, an MCTS search is employed where a policy network narrows down move selection (i.e., effectively reducing the branching factor) and a value network helps with leaf evaluation (i.e., reducing the number of costly rollouts to estimate state-value of leaf nodes).

Our work differs from AlphaGo-like works in multiple aspects: (i) our framework specifically aims to enable on-policy model-free RL methods to explore \emph{safely} in hard-exploration domains where negative rewarding terminal states are ubiquitous, in contrast to off-policy methods which use intensive search to fully learn a policy; (ii) our framework is general in that other demonstrators (human or other sources) can be integrated to provide action guidance by using the proposed auxiliary loss refinement; and
(iii) our framework aims to use the demonstrator with a small lookahead (i.e., shallow) search to filter out actions leading to immediate negative terminal states so that model-free RL can imitate those safer actions to learn to safely explore.

\section{Preliminaries}  

\subsection{Reinforcement Learning}

We assume the standard reinforcement learning setting of an agent interacting in an environment over a discrete number of steps. At time $t$ the agent in state $s_t$ takes an action $a_t$ and receives a reward $r_t$. The discounted return is defined as $R_{t:\infty} = \sum_{t=0}^\infty \gamma^t r_t$. The state-value function, $V^\pi(s)=\mathbb{E}[R_{t:\infty}|s_t=s,\pi]$ is the expected return from state $s$ following a policy $\pi(a|s)$
and the action-value function, $Q^\pi(s,a)=\mathbb{E}[R_{t:\infty}|s_t=s, a_t=a,\pi]$, is the expected return following policy $\pi$ after taking action $a$ from state $s$.

The A3C method, as an actor-critic algorithm, has a policy network (actor) and a value network (critic) where the actor is parameterized by $\pi(a|s;\theta)$, updated by ${\triangle\theta = \nabla_\theta \log \pi(a_t|s_t; \theta) A(s_t, a_t; \theta_v)}$, and the critic is parameterized by $V(s; \theta_v)$, updated by ${\triangle\theta_v = A(s_t, a_t; \theta_v) \nabla_{\theta_v} V(s_t)}$ where $$A(s_t, a_t; \theta_v) = \sum_k^{n-1} \gamma^kr_{t+k} + \gamma^n V(s_{t+n}) - V (s_t)$$
with $A(s,a)=Q(s,a)-V(s)$ representing the \emph{advantage} function.

The policy and the value function are updated after every $t_{max}$ actions or when a terminal state is reached. It is common to use one softmax output layer for the policy $\pi(a_t|s_t; \theta)$ head and one linear output for the value function $V (s_t; \theta_v)$ head, with all non-output layers shared.

The loss function for A3C is composed of two terms: policy loss (actor), $\mathcal{L}_{\pi}$, and value loss (critic), $\mathcal{L}_{v}$. An entropy loss for the policy, $H(\pi)$, is also commonly added, which helps to improve exploration by discouraging premature convergence to suboptimal deterministic policies~\cite{mnih2016asynchronous}. Thus, the loss function is given by: $$\mathcal{L}_{\text{A3C}} = \lambda_v  \mathcal{L}_{v} + \lambda_{\pi} \mathcal{L}_{\pi} - \lambda_{H} \mathbb{E}_{s \sim \pi} [H(\pi(s, \cdot, \theta)] $$ with $\lambda_{v}=0.5$, $\lambda_{\pi}=1.0$, and $\lambda_{H}=0.01$, being standard weighting terms on the individual loss components.

UNREAL~\cite{jaderberg2016reinforcement}, which is based on A3C, proposed unsupervised \emph{auxiliary tasks} (e.g., reward prediction task where the agent predicts the next reward, which is observed in a self-supervised fashion) to speed up learning. In contrast to vanilla A3C, UNREAL uses an experience replay buffer that is sampled with more priority given to positively rewarded interactions to improve the critic network, and it uses off-policy learning. We compare against A3C rather than UNREAL for better ablation by just augmenting the auxiliary task on top of A3C.

\subsection{Monte Carlo Tree Search} 

MCTS is a best-first search algorithm that gained traction after its breakthrough performance in Go \cite{coulom2006efficient}. It has been used for many purposes, e.g., for game playing~\cite{sturtevant2015monte,silver2016mastering}, for playtesting~\cite{zook2015monte,holmgaard2018automated,borovikov2019winning}, and for robotics~\cite{kartal2015stochastic,best2019dec}.

In MCTS, a search tree is generated where each node in the tree represents a complete state of the domain and each link represents one possible valid action, leading to a child node representing the resulting state after taking an action. The root of the tree is the initial state (for example, the initial configuration of the Pommerman board including the agent location).
MCTS proceeds in four phases of: selection, expansion, rollout, and back-propagation. The standard MCTS algorithm proceeds by repeatedly adding one node at a time to the current tree. Given that leaf nodes are likely to be far from terminal states, it uses random actions, a.k.a. \textit{rollouts}, to estimate state-action values. The rollout policy can also be biased based on information obtained during search, or external domain knowledge. After the rollout phase, the total collected rewards during the episode is back-propagated through the tree branch, updating their empirical state-action values, and visit counts.

\paragraph{Exploration vs.~Exploitation Dilemma}

Choosing which child node to expand (i.e., choosing an action) becomes an exploration/exploitation problem given the empirical estimates. Upper Confidence Bounds (UCB1)~\cite{auer2002finite} is a bandit algorithm that is used for such settings with provable guarantees. Using UCB1 with MCTS is also referred to as Upper Confidence bounds applied to Trees (UCT)~\cite{kocsis2006bandit}. Applied to our framework, each parent node $s$ chooses its child $s'$ with the largest $UCB1(s,a)$ value according to Eqn.~\ref{eqn:ucb}.

\begin{equation}
UCB1(s,a) = Q(s,a) + C \sqrt{\frac{\ln n(s)}{n(s')}}
\label{eqn:ucb}
\end{equation}

\noindent Here, $n(s)$ denotes number of visits to the node $s$ and $n(s')$ denotes the number of visits to $s'$, i.e., the resulting child node when taking action $a$ from node $s$. The value of $C$ is a tunable exploration parameter.

\section{Safer RL with MCTS Action Guidance}

\begin{figure*}
\centering
\includegraphics[width=\linewidth]{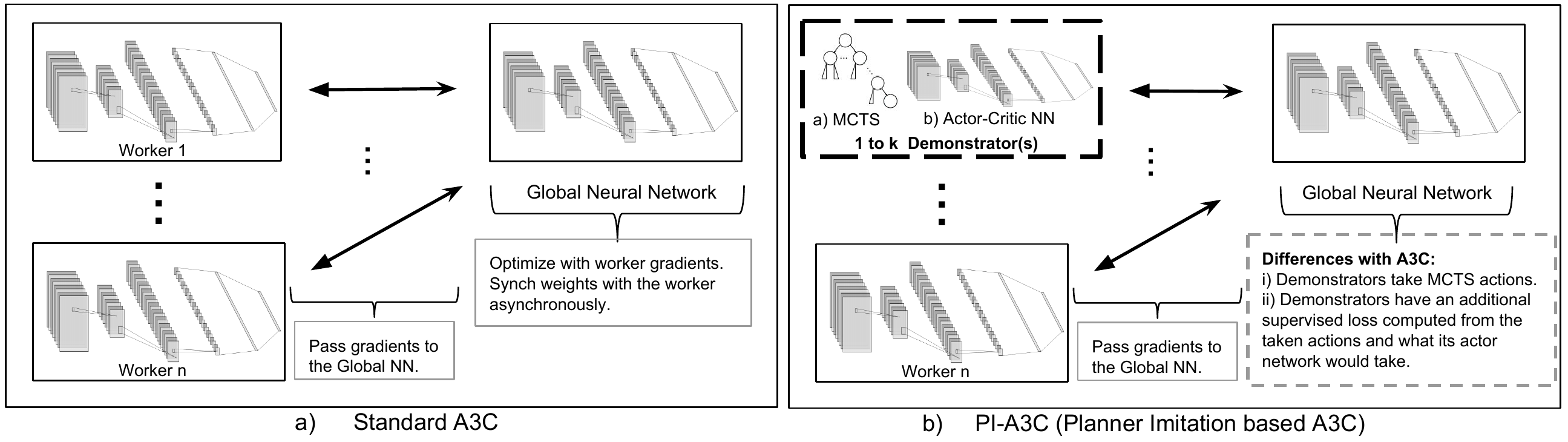}
\caption{\textbf{a)} In the A3C framework, each worker independently interacts with the environment and computes gradients. Then, each worker \emph{asynchronously} passes the gradients to the global neural network which updates parameters and synchronizes with the respective worker. \textbf{b)} In Planner Imitation based A3C (PI-A3C), $k\ge 1$ CPU workers are assigned as demonstrators taking MCTS actions, while keeping track of what action its actor network would take. The demonstrator workers have an additional auxiliary supervised loss. PI-A3C enables the network to simultaneously optimize the policy and learn to imitate MCTS.}
\label{fig:pi_a3c}
\end{figure*}

In this section, we present our framework PI-A3C (Planner Imitation - A3C) that extends A3C with a lightweight search-based demonstrator through an auxiliary task based loss refinement as depicted in Figure~\ref{fig:pi_a3c}. %

We propose a framework that can use planners, or other sources of demonstrators, within asynchronous DRL methods to accelerate learning.
Even though our framework can be generalized to a variety of planners and DRL methods, we showcase our contribution using MCTS and A3C. In particular, we employ the UCB1 method to balance exploration versus exploitation during planning. During rollouts, we simulate all agents as random agents as in vanilla MCTS and we perform limited-depth rollouts for time-constraints. 

The motivation for combining MCTS and asynchronous DRL methods stems from the need to decrease training time, even if the planner or the simulator is very slow. In this work, we assume the demonstrator and actor-critic networks are decoupled, i.e., a vanilla UCT planner is used as a black-box that takes an observation and returns an action resembling UCTtoClassification method~\cite{guo2014deep}, which uses slower MCTS to generate an expert move dataset and trains a NN in a supervised fashion to imitate the search algorithm.

However, the UCTtoClassification method used a high-number of rollouts (10K) per action selection to construct an expert dataset. In contrast, we show how vanilla MCTS with a small number of rollouts\footnote{In Pommerman 10K rollouts would take hours due to very slow simulator and long horizon~\cite{matiisen2018pommerman}.} ($\approx$ 100) can still be employed in an \textit{on-policy} fashion to improve training efficiency for actor-critic RL in challenging domains with abundant, easily reachable, terminal states with negative rewards.

Within A3C's asynchronous distributed architecture, all the CPU workers perform agent-environment interactions with their policy networks, see Figure~\ref{fig:pi_a3c}(a). In our new framework, PI-A3C (Planner Imitation with A3C), we assign $k\ge 1$ CPU 
workers (we performed experiments with different $k$ values) to perform MCTS-based planning for agent-environment interaction based on the agent's observations, while also keeping track of how its policy network would perform for those cases, see Figure~\ref{fig:pi_a3c}(b). In this fashion, we both learn to imitate the MCTS planner and to optimize the policy. The main motivation for the PI-A3C framework is to increase the number of agent-environment interactions with positive rewards for hard-exploration RL problems to improve training efficiency.

Note that the planner-based worker still has its own neural network with actor and policy heads, but action selection is performed by the planner while its policy head is used for loss computation. In particular, the MCTS planner based worker augments its loss function with the auxiliary task of \emph{Planner Imitation}\footnote{Both Guo et al.~(\citeyear{guo2014deep}) and Anthony et al.~(\citeyear{anthony2017thinking}) used MCTS moves as a learning target, referred to as \textit{Chosen Action Target}.}. The auxiliary loss is defined as ${\mathcal{L}_{PI}= -\frac{1}{N} \sum_i^N a^i \log (\hat{a}^i) }$,  which is the supervised cross entropy loss between  $a^i$ and $\hat{a}^i$, representing the one-hot encoded action the planner used and the action the policy network would take for the same observation respectively during an episode of length $N$. 
 The demonstrator worker's loss after the addition of \emph{Planner Imitation} is defined by $$\mathcal{L}_{\text{PI-A3C}}= \mathcal{L}_{A3C} + \lambda_{PI} \mathcal{L}_{PI} $$ 
\noindent where $\lambda_{PI}=1$ is a weight term (which was not tuned). 
In PI-A3C non-demonstrator workers are left unchanged. By formulating the \textit{Planner Imitation} loss as an auxiliary loss, the objective of the resulting framework becomes a multi-task learning problem where the agent learns to both maximize the reward and imitate the planner.

\section{Experiments and Results}

In this section, we present the experimental setup and results in the simplified version of the Pommerman game.

\subsection{Pommerman}

In Pommerman, each agent can execute one of 6 actions at every timestep: move in any of four directions, stay put, or place a bomb. Each cell on the board can be a passage, a rigid wall, or wood. The maps are generated randomly, albeit there is always a guaranteed path between agents.
Whenever an agent places a bomb it explodes after 10 timesteps, producing flames that have a lifetime of 2 timesteps. Flames destroy wood and kill any agents within their blast radius. %
An episode can last up to 800 timesteps. Pommerman is challenging due to the following characteristics:

\textbf{Multiagent component:} the agent needs to best respond to any type of opponent, but agents' behaviours also change based on collected power-ups.

\textbf{Delayed action effects}: the only way to make a change to the environment (e.g., kill an agent) is by means of bomb placement, but the effect of such an action is only observed when the bombs explodes after 10 time steps.

\textbf{Sparse and deceptive rewards}: the former refers to the fact that rewards are episodic. The latter refers to the fact that quite often a winning reward is due to the opponents' involuntary \emph{suicide}, which makes reinforcing an agent's action based on such a reward \emph{deceptive}. 

For these reasons, we consider this game challenging for many standard RL algorithms and a local optimum is commonly learned, i.e., not placing bombs~\cite{resnick2018pommerman}.

\subsection{Setup}

 We run ablation experiments on the contribution of \emph{Planner Imitation} to compare against the standard A3C method for: (i) single demonstrator with different expertise levels, (ii) different number of demonstrators with the same expertise level, and (iii) using rollout biasing within MCTS in contrast to uniform random rollout policy.
 
We considered two types of opponents:

\begin{itemize}
    \item \emph{Static} opponents: the opponent waits in the initial position and always executes the `stay put' action. This opponent provides the easiest configuration (excluding suicidal opponents). It is a baseline opponent to show how challenging the game is for model-free RL.
    
    \item \emph{Rule-based} opponents: this is the benchmark agent within the simulator. It collects power-ups and places bombs when it is near an opponent. It is skilled in avoiding blasts from bombs. It uses Dijkstra's algorithm on each time-step, resulting in longer training times.
\end{itemize}

All training curves are obtained from 3 runs with different random seeds. From these 3 separate runs, we compute average learning performance (depicted with bold color curves) and standard deviations (depicted with shaded color curves). At every training episode, the board is shuffled.

Because of all the complexities mentioned in this domain we simplified it by considering a game with only two agents and a reduced board size of $8 \times 8$, see Figure~\ref{fig:pom8x8}. Note that we still randomize \emph{the location of} walls, wood, power-ups, and the initial position of the agents for every episode.

\begin{figure*}
    \subfloat[Learning against a Static Agent]{{\includegraphics[scale=0.2975]{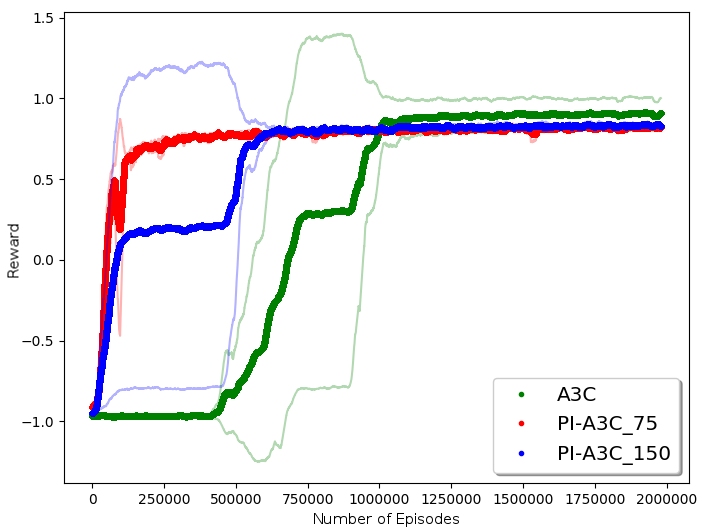} }}
    \subfloat[Learning against a Rule-based Agent]{{\includegraphics[scale=0.2675]{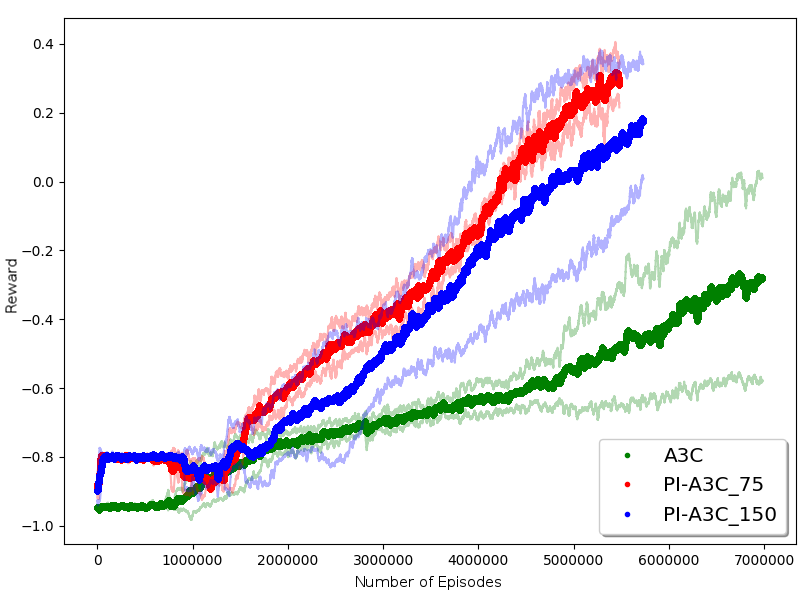} }}%
    \subfloat[Varying the number of Demonstrators]{{\includegraphics[scale=0.2175]{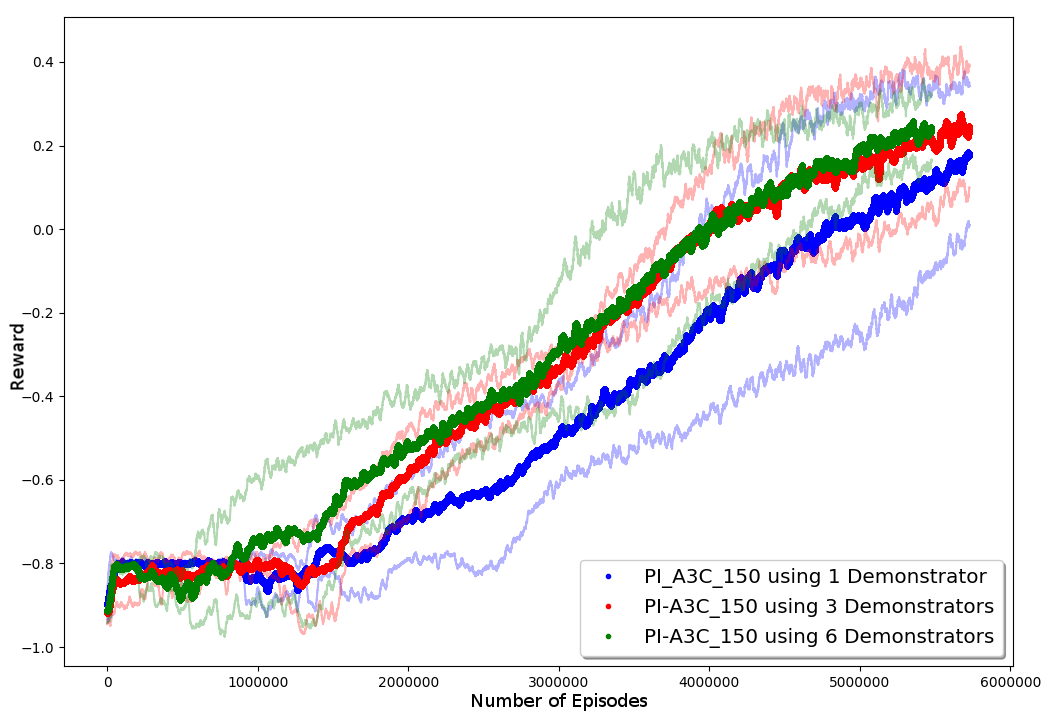} }}%

    \caption{ a) Against Static agent, all variants have been trained for 12 hours. The ${\text{PI-A3C}}$ framework using MCTS demonstrator with 75 and 150 rollouts learns significantly faster compared to the standard A3C. b) Against Rule-based opponent, all variants have been trained for 3 days. Against this more skilled opponent, PI-A3C provides significant speed up in learning performance, and finds better best response policies. c) Against Rule-based opponent, employing different number $(n=1,3,6)$ of Demonstrators, i.e. increasing from 1 to 3, slightly improved the results, however, there is almost no variation from 3 to 6 demonstrators. For both (a) and (b), increasing the expertise level of MCTS through doubling the number of rollouts (from 75 to 150) does not yield improvement, and can even hurt performance. Our hypothesis is that slower planning decreases the number of demonstrator actions too much for the model-free RL workers to learn to imitate for safe exploration.}%
    \label{fig:ablate_planner}%
\end{figure*}

\subsection{Results}
\label{sec:results}

We conducted two sets of experiments learning against \emph{Static} and \emph{Rule-based} opponents. We benchmark against the standard A3C with learning curves in terms of converged policies and time-efficiency. All approaches were trained using 24 CPU cores. Training curves for PI-A3C variants are obtained by using only the model-free workers, excluding rewards obtained by the demonstrator worker, to accurately observe the performance of model-free RL.

\paragraph{On action guidance, quantity versus quality}

Within our framework, we can vary the expertise level of MCTS by simply changing the number of rollouts. We experimented with 75 and 150 rollouts. Given finite training time, higher rollouts imply deeper search and better moves, however, it also implies that number of guided actions by the demonstrators will be fewer in quantity, reducing the number of asynchronous updates to the global neural network. As Figure~\ref{fig:ablate_planner} shows, the relatively weaker demonstrator (75 rollouts) enabled faster learning than the one with 150 rollouts against both opponents.
We also measured the Vanilla-MCTS performance against both opponents as reported in Table~\ref{tab:table_mcts}. The results show that average rewards increase when given more rollouts against both opponents. Against static opponents, Vanilla-MCTS performance is worse than both A3C and PI-A3C methods. This occurs because even when vanilla-MCTS does not commit suicides often, it fails to blast the opponent, resulting in too many tie games (episodic reward of -1). In contrast, PI-A3C makes use of the demonstrator for safer exploration with higher-quality atomic actions and learns to explore faster than A3C.
 Against a more challenging Rule-based opponent, pure MCTS (mean episodic reward of $-0.23$ using 75 rollouts) is slightly better than pure A3C (mean episodic reward of $-0.27$), however PI-A3C with either 75 or 150 rollouts combines both approaches well and perform the best, see Figure~\ref{fig:ablate_planner}(b).

\begin{table}
\footnotesize
    \centering
        \caption{Vanilla MCTS-based demonstrator evaluation: average episodic rewards of 200 games. Note that mean rewards are affected by too many tie games (rewarded -1 by the Pommerman simulator), e.g. against Static opponent, MCTS wins 88 games, loses 4 games by suicide, and ties 108 games, which results in average reward of -0.12.}
    \begin{tabular}{c|c |c }
        \toprule
        Vanilla-MCTS vs. & \texttt{Static} & \texttt{Rule-based} \\ 
        \midrule
         75 Rollouts & -0.12 & -0.23  \\
         150 Rollouts & -0.08 & -0.20 \\
         \bottomrule
    \end{tabular}
    \label{tab:table_mcts}
\end{table}

We hypothesize that the faster demonstrator (MCTS with only 75 rollouts) makes more updates to the global neural network, warming up other purely model-free workers for \emph{safer} exploration earlier in contrast to the slower demonstrator. This is reasonable as the model-free RL workers constitute all but one of the CPU workers in these experiments, therefore the earlier the model-free workers can start safer exploration, the better the learning progress is likely to be.\footnote{Even though we used MCTS with fixed number of rollouts, this could be set dynamically, for example, by exploiting the reward sparsity or variance specific to the problem domain, e.g., using higher number of rollouts when close to bombs or opponents.}

\paragraph{On the trade-off of using multiple demonstrators}

Our proposed method is built on top of an asynchronous distributed framework that uses several CPU workers: $k\ge 1$ act as a demonstrator and the rest of them explore in model-free fashion. We conducted one  experiment to better understand how increasing the number of demonstrators, each of which provides additional \emph{Planner Imitation} losses asynchronously, affects the learning performance. The trade-off is that more demonstrators imply fewer model-free workers to optimize the main task, but also a higher number of actions to imitate. We present the results in Figure~\ref{fig:ablate_planner}(c) where we experimented 3 and 6 demonstrators, with identical resources and with 150 rollouts each. Results show that increasing to 3 improves the performance while extending to 6 demonstrators does not provide any marginal improvement. We also observe that the 3 demonstrator version using 150 rollouts presented in Figure~\ref{fig:ablate_planner}(c) has a relatively similar performance with the 1 demonstrator version using 75 rollouts (see Figure~\ref{fig:ablate_planner}(b)), which is aligned with our hypothesis that providing more demonstrator guidance early during learning is more valuable than fewer higher quality demonstrations. 
\paragraph{Demonstrator biasing with policy network}

\begin{figure}
\centering
\includegraphics[scale=0.305]{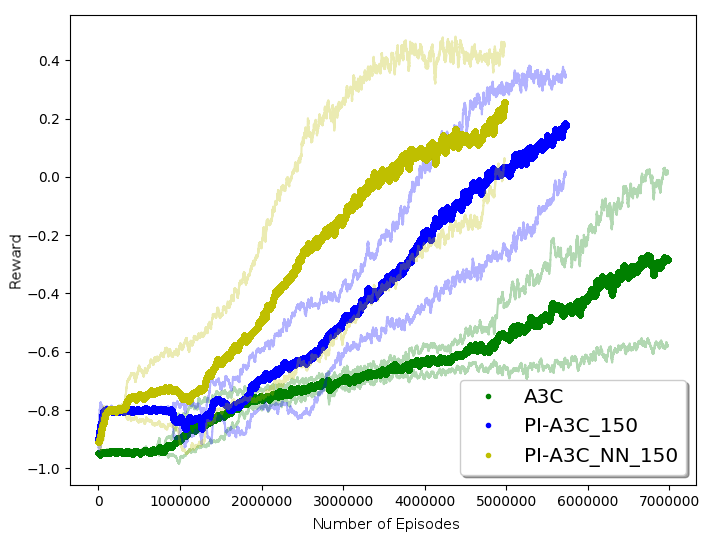}
\caption{Learning against Rule-based opponent: using the policy head during MCTS rollout phase within the demonstrator provides improvement in learning speed, but it has higher variance compared to the default random policy.}
\label{fig:simple_rollout_biasing}
\end{figure}

Uniform random rollouts employed by MCTS yield unbiased estimates, however it requires a high number of rollouts to reduce the variance. One way to improve search efficiency has been through different biasing strategies, such as prioritizing actions globally based on their evaluation scores~\cite{kartal2014user}, 
using heuristically computed move urgency values~\cite{bouzy2005associating}, or concurrently learning a rollout policy~\cite{ilhan2017monte}. In a similar vein with these methods, we let the MCTS-based demonstrator to access the policy head during rollouts, named as PI-A3C-NN (Planner Imitation - A3C with Neural Network). Our results suggest that employing a biased rollout policy improves in the average learning performance, however it has higher variance, as depicted in Figure~\ref{fig:simple_rollout_biasing}.

\section{Conclusions} 

In this work, we present a framework that uses a non-expert simulated demonstrator within a distributed asynchronous deep RL method to succeed in hard-exploration domains. Our experiments use the recently proposed Pommerman domain, a very challenging benchmark for pure model-free RL methods. In Pommerman, one main challenge is the high probability of suicide while exploring (yielding a negative reward), which occurs due to the delayed bomb explosion. However, the agent cannot succeed without learning how to stay safe after bomb placement. The methods and ideas proposed in this paper address this hard-exploration challenge. The main idea of our proposed framework is to use MCTS as a shallow demonstrator (small number of rollouts). In our framework, model-free workers learn to safely explore and acquire this skill by imitating a shallow search-based non-expert demonstrator. We performed different experiments varying the quality and the number of demonstrators. The results show that our proposed method shows significant improvement in learning efficiency across different opponents.

There are several directions to extend our work. Soemers et al.~(\citeyear{soemers2016enhancements}) employed Breadth-First Search to initialize state-action value estimates by identifying near terminal states before starting MCTS based lookahead search. This enhancement can be tested within our framework. Also, MCTS-minimax hybrids method~\cite{baier2018mcts} can be employed as the demonstrator, which can provide better tactical play through minimax search. Another avenue is to investigate how to formulate the problem so that \emph{ad-hoc} invocation of an MCTS-based simulator can be employed, i.e., action guidance can be employed only when the model-free RL agent needs one, e.g., in Pommerman this means whenever a bomb is about to go off or near enemies where enemy engagement requires strategic actions. 

All of our work presented in the paper is on-policy for better comparison to the standard A3C method --- we maintain no experience replay buffer. This means that MCTS actions are used only once to update neural network and thrown away. In contrast, UNREAL uses a buffer and gives higher priority to samples with positive rewards. We could take a similar approach to save demonstrator's experiences to a buffer and sample based on the rewards.

\section*{Appendix}

\textbf{Neural Network Architecture:} We use a NN with 4 convolutional layers that have 32 filters and $3 \times 3$ kernels, with stride and padding of 1, followed with a dense layer with 128 units, followed actor-critic heads. Architecture is not tuned.

\textbf{State Representation:} Similar to ~\cite{resnick2018pommerman}, we maintain 28 feature maps that are constructed from the agent observation. These channels maintain location of walls, wood, power-ups, agents, bombs, and flames. Agents have different properties such as bomb kick, bomb blast radius, and number of bombs. We maintain 3 feature maps for these abilities per agent, in total 12 is used to support up to 4 agents. We also maintain a feature map for the remaining lifetime of flames. All the feature channels can be readily extracted from agent observation except the opponents' properties and the flames' remaining lifetime, which are tracked efficiently from sequential observations.

\textbf{Hyperparameter Tuning:} We did not perform a through hyperparameter tuning due to long training times. We used a $\gamma=0.999$ for discount factor. For A3C loss, the default weights are employed, i.e., $\lambda_{v}=0.5$, $\lambda_{\pi}=1.0$, and $\lambda_{H}=0.01$. For the \emph{Planner Imitation} task, $\lambda_{PI}=1$ is used for the MCTS worker. We employed the Adam optimizer with a learning rate of $0.0001$. We found that for the Adam optimizer, $\epsilon = 1\times10^{-5}$ provides a more stable learning curve (less catastrophic forgetting). We used a weight decay of $1\times10^{-5}$ within the Adam optimizer for L2 regularization.

\textbf{MCTS Implementation Details:} After rollouts, demonstrator action is selected by the max visit count, a.k.a. \textit{Robust Child}~\cite{browne2012survey}. Default UCB1 exploration constant of $\sqrt{2}$ is used. The max search tree depth from any given state is set to 25 given the low search budget. The value head by the NN is not used by vanilla MCTS, %

\bibliographystyle{aaai}
\clearpage
\small
\bibliography{ref}

\end{document}